\documentclass{article}

\usepackage{PRIMEarxiv}

\usepackage[utf8]{inputenc} 
\usepackage[T1]{fontenc}    
\usepackage{hyperref}       
\usepackage{url}            
\usepackage{booktabs}       
\usepackage{amsfonts}       
\usepackage{nicefrac}       
\usepackage{microtype}      
\usepackage{lipsum}
\usepackage{fancyhdr}       
\usepackage{graphicx}       
\graphicspath{{media/}}     

\title{A Performance Evaluation of a Quantized Large Language Model 
on Various Smartphones
}

\author{
  Tolga Çöplü, Marc Loedi, Arto Bendiken, Mykhailo Makohin, Joshua J. Bouw, Stephen Cobb,\\
  Haltia, Inc. \\
  \texttt{\{tolga, marc, arto, mykhaylo, joshua, steve\}@haltia.ai} \\
}

\begin{document}
\maketitle

\begin{abstract}
This paper explores the feasibility and performance of on-device large language model (LLM) inference on various Apple iPhone models. Amidst the rapid evolution of generative AI, on-device LLMs offer solutions to privacy, security, and connectivity challenges inherent in cloud-based models. Leveraging existing literature on running multi-billion parameter LLMs on resource-limited devices, our study examines the thermal effects and interaction speeds of a high-performing LLM across different smartphone generations. We present real-world performance results, providing insights into on-device inference capabilities. 

\end{abstract}


\section{Introduction}
Today's popular generative AI applications, such as ChatGPT and GitHub Copilot, operate on cloud-based systems. The cloud model’s primary value proposition is faster inference by running memory and processing-intensive large language models (LLM) on powerful and scalable cloud servers. Although this approach offers major benefits for certain AI applications, the inevitability of on-device LLMs becomes apparent when considering applications where data ownership, safety, and security are crucial. It is also of great importance for the democratization of the AI domain that users run applications on their own devices. Without a paradigm shift in AI technologies—similar to the transition from mainframes to personal computers, and subsequently to mobile devices—we will be unable to unlock AI's true potential. Fortunately, remarkable developments in the LLM field are emerging every day. Thanks to the R\&D efforts in LLM architecture, LLMs with fewer parameters are now offering faster responses and improved performance compared to older large models. We have indeed reached a stage where LLM quality takes precedence over quantity. The number of models demonstrating performance superior to their larger counterparts is increasing daily \cite{open-llm-leaderboard}.

Running LLMs on smartphones or even other edge devices has significant advantages. Most importantly, an on-device LLM provides privacy and security \cite{marin_serverless_2022}. Transmitting personal data entered into health applications or private conversations with a personal assistant over the internet is a vulnerability. Disconnection is another important issue that cannot be addressed in cloud-based architecture. It is unacceptable for AI applications, which we will utilize in critical moments of our daily lives, to become even temporarily unusable due to connectivity issues. Also, transmission latency becomes a limiting factor for time-sensitive applications. All things considered, we anticipate different approaches to the generative AI and LLM revolution, including cloud-based, on-device, and hybrid architectures.

In this study, we will be focusing on the feasibility of running LLMs on smartphones. We will investigate the question “Can a state-of-the-art LLM with billions of parameters truly operate on a smartphone without any network connectivity?” We will concentrate on the interaction speed of a performant LLM on various iPhones, and we will evaluate on-device inference capabilities by empirical data gathered through our tests. 

The structure of this paper is as follows: in Section 2, we examine the proposed methods for running state-of-the-art LLMs on devices with limited resources. Section 3 describes our experimental setup and test procedures. Section 4 presents our test results and their interpretations. Section 5 concludes the paper and outlines future directions.

\section{Related Work}
Smartphones are resource-constrained devices for running today’s multi-billion-parameter LLMs. In particular, LLMs’ memory footprint makes it infeasible to run most LLMs natively even on top-of-the-line, maximally configured smartphones. Fortunately, rapidly progressing AI research offers hope. The literature suggests many valuable approaches for compressing model sizes. Among these methodologies, Knowledge Distillation, Pruning, and Quantization emerge as prominent strategies.

\subsection{Knowledge Distillation}
Knowledge distillation (KD) \cite{hinton_distilling_2015, kim_sequence-level_2016} is a machine learning technique employed to reduce LLM size and complexity to shorten inference times. This method is designed for the transfer of refined and compressed knowledge from a sophisticated model, named “teacher model”, to what is designated as the “student model”. As with all compression techniques, the goal is to achieve maximum compression with minimum quality loss. Although some KD techniques have achieved substantial success \cite{huang_-context_2022, magister_teaching_2023, gu_knowledge_2023}, integration challenges made KD infeasible for model compression within the scope of this study.

\subsection{Pruning}
Pruning can be defined as the reduction in size and processing requirements of an LLM by removing redundant or ineffective elements (weights, neurons, and/or layers) that vary according to the method. Not only can pruning notably reduce LLM size, but the removal of redundant/inefficient elements also results in an acceleration of the inference mechanism. A critical point to consider regarding this method is whether training is required either during or after pruning \cite{zhu_survey_2023}. Cutting-edge pruning methodologies, such as SparseGTP \cite{frantar_sparsegpt_2023} and Wanda\cite{sun_simple_2023}, can achieve success even at high sparsity without the need for re-training. Due to integration challenges, similar to KD, pruning techniques were not employed in our current study.

\subsection{Quantization}
In the LLM domain, quantization can be defined as the process of reducing the bit-width of model parameters. LLMs typically store weight parameters as 32-bit floating points. Quantization attempts to maintain these with lower resolution with minimal quality loss. In the literature, LLM-specific quantization has been extensively studied for various resolutions, such as 8-bit and 4-bit integers \cite{noune_8-bit_2022, dettmers_llmint8_2022, shao_omniquant_2023}. Additionally, there exist distinct methodologies in the literature either focusing on the execution of quantization during the training phase, or quantizing the parameters of a trained model \cite{zhu_survey_2023}. Depending on the quantization method employed, an LLM’s memory footprint and processing requirements can be significantly reduced. However, it is essential to note that quantization introduces a precision loss, which affects inference performance. The evaluation of inference degradation due to quantization is a research area we wish to elaborate on in future work. Nonetheless, we would like to highlight that our preliminary benchmarks have yielded promising results. For our study, owing to accessibility and ease of integration, we have opted to conduct our tests using a quantized model.

\section{Experimental Setup}
The primary objective of our experiments is to evaluate sampling, prompt decoding, and inference speeds on various widely used iPhone models. Our aim is twofold: first, to evaluate the execution speed of a state-of-the-art LLM on different smartphones to see if the results are acceptable, and second, to see how on-device LLM performance is affected in evolving smartphone generations. 

We established a test bed consisting of five iPhone models: 14, 14 Pro, 14 Pro Max, 15, and 15 Pro Max, all of which have at least 6 GiB of RAM. All models used in the test bed are equipped with a GPU. Since 2014, the GPU has been optimally supported in iOS via the Metal GPU shader API. The benchmarking suite we developed runs inference on the GPU using Metal and sampling on the CPU. For the results to be consistent and comparable, the tests were conducted on smartphones with 100\% battery charge, with no background applications running, and using the same iOS version (iOS 17.1.x).

In selecting the LLM for the tests, the Hugging Face Leadership benchmark \cite{open-llm-leaderboard} results were considered. Because of its clean inference and consistent verbosity, the orca\_mini\_v3\_7B \cite{mathur2023orcamini} model, which uses the Llama 2 model \cite{touvron2023llama} architecture but trained on the Orca Style dataset \cite{mukherjee2023orca}, was found suitable. Additionally, support for different quantization resolutions in the GGUF file format is available to accommodate the model's memory footprint on smartphones. We would like to extend our gratitude to the llama.cpp \cite{llama.cpp} project for providing the GGUF format in terms of both performance and usability. In this study, the 3-bit small K-S method was employed. The size of the model used is 2.95 GiB.

\section{Performance Evaluation}
The resource usage of mobile applications, such as CPU, GPU, and I/O, is reduced by the operating system based on the application's energy impact. On iOS, this mechanism is managed through four different thermal state levels as given below.
\begin{itemize}
\item Nominal
\item Fair
\item Serious
\item Critical
\end{itemize}

\begin{figure}[ht]
  \centering
  \includegraphics[scale=1.1]{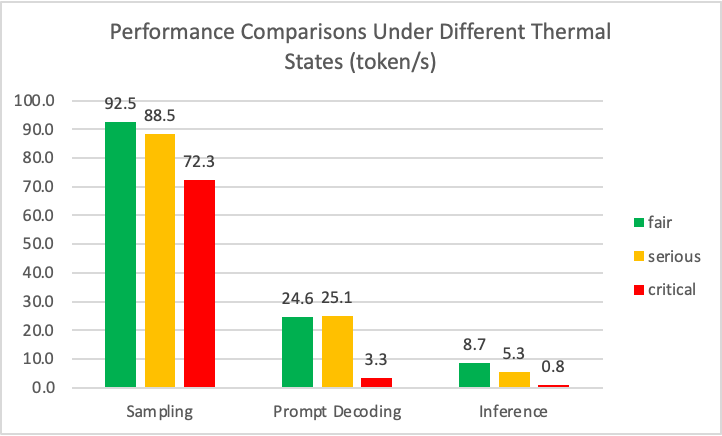}
  \caption{Sampling, prompt decoding, and inference results under different thermal states.}
  \label{fig:fig1}
\end{figure}

In our test suite, sampling CPU, prompt decoding, and inference operations are GPU-intensive tasks. Therefore, we will first examine how much the thermal state affects our test results. In tests conducted via Xcode, devices can be conditioned to “fair”, “serious”, or “critical” thermal states. Using the iPhone 14 Pro, our test suite was executed under these conditioned thermal states, and the results are provided in Figure.\ref{fig:fig1}.

As can be seen from Figure.\ref{fig:fig1}, in the critical state, the application's performance remains below acceptable limits compared to the other states. In fair and serious thermal states, although there are no major differences in the sampling and prompt decoding tests, a significant difference is observed in the inference test.

\begin{figure}[ht]
  \centering
  \includegraphics[scale=1.1]{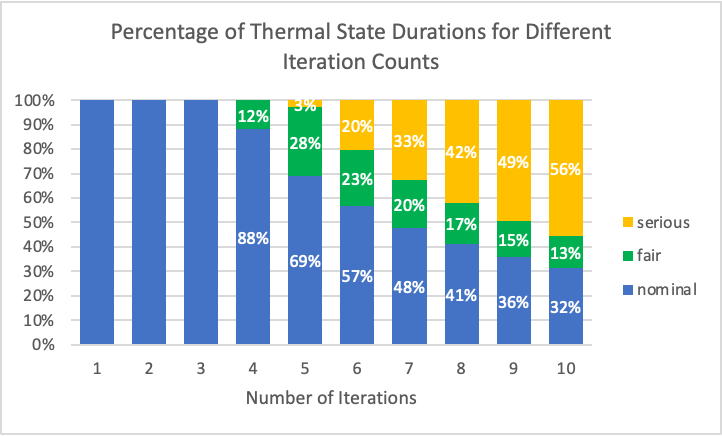}
  \caption{Thermal state duration ratios for different iteration counts.}
  \label{fig:fig2}
\end{figure}

We can control the energy impact of our test suite based on two parameters: the number of iterations determining how long the application will run, and the delay between iterations. Since delay is a sleep cycle for the application, the application's energy impact decreases. In our tests, it has been observed that the thermal state does not exceed the fair range regardless of the number of iterations when a delay of 90 seconds is provided between prompts. However, considering the delay as the duration for the user to prepare input to the LLM, a 90-second delay is not practical from the perspective of a real usage scenario. For a more realistic scenario, our tests will be run with a five-second delay. The relationship between the number of iterations and the thermal state has been examined on an iPhone 14 Pro in Figure.\ref{fig:fig2}. According to Figure.\ref{fig:fig2}, the thermal state of the smartphone does not exceed the nominal range for up to three iterations. Larger iteration counts result in observations of fair and serious thermal states.

To isolate results from the effects of high thermal states, we conducted all subsequent tests using the following configurations:
\begin{itemize}
\item Each test was run three times on each device with three iterations and the results were presented based on the average of the nine measurements. 
\item There was a five-second delay between iterations.
\item For each test iteration, the entire context was reset, and the model was prompted with the question, "Why is the sky blue?".
\item To isolate the test results from delays caused by factors such as initialization, late binding, etc, a proper one-time warm-up was applied.
\item Eight threads were used to schedule the Metal shaders.
\item The LLM’s context length and batch length configurations were set to 1024 and 512, respectively.
\item The sampling approach employed is “greedy sampling”, where the most probable token is consistently chosen as the next in the sequence, resulting in deterministic text generation.
\end{itemize}

\subsection{Sampling Performance}
The selected LLM’s sampling performance for varying iPhone models was evaluated based on the number of tokens sampled per second. Sampling refers to picking the next token based on the probability distribution over the LLM’s vocabulary. Our goal being to investigate the throughput of the same sampling method on different smartphones, different sampling methods were not investigated, and the tests were run with the greedy sampling. Figure.\ref{fig:fig3} presents the token sampling throughput across different iPhone models. 

Sampling is executed on the CPU. As can be seen from Figure.\ref{fig:fig3}, the iPhone 14 Pro, 14 Pro Max, and 15 models, which all have the same 6-core CPU A16 Bionic chip, have produced similar results. The iPhone 15 Pro Max model, using the A17 Pro chip with a 6-core CPU, shows about 20\% higher performance in tests compared to other devices. Our test results indicate that although the sampling throughput in the iPhone 14 model, which uses the 6-core CPU A15 Bionic, is slightly lower, there is no significant difference.

\subsection{Prompt Decoding Performance}
Prompt decoding performance is indicative of the speed at which the prompt input is processed. At a high level, prompt decoding is creating an embedding matrix for the input tokens and performing the rest of the matrix multiplications defined in the model's architecture to compute the logits corresponding to the input, which are the non-normalized raw vectors corresponding to what the LLM predicts the next token to likely be. For each and every test scenario, the prompt input remains consistent: “Why is the sky blue?”. The performance of the model’s prompt processing for varying iPhone models is again evaluated based on the number of tokens processed per second. 

\begin{figure}[ht]
  \centering
  \includegraphics[scale=0.75]{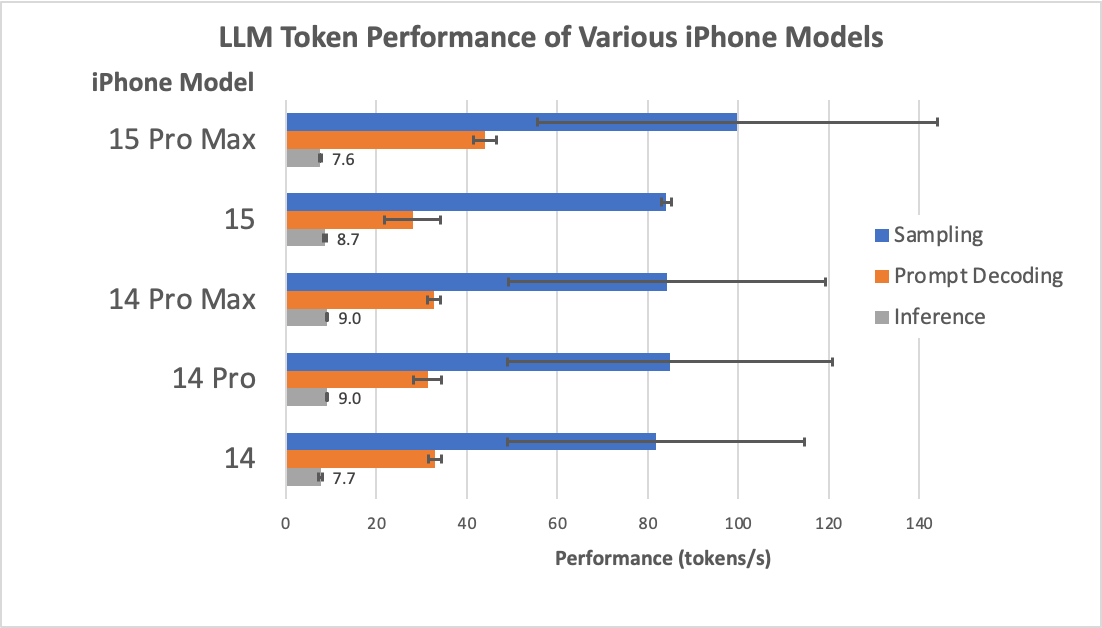}
  \caption{Sampling, prompt decoding, and inference token rates of various iPhone models.}
  \label{fig:fig3}
\end{figure}

Prompt decoding runs on the GPU. As can be seen from Figure.\ref{fig:fig3}, devices using the A16 Bionic chipset with the same 5-core GPU (iPhone 14 Pro, 14 Pro Max, and 15) have shown different results this time. The iPhone 14 Pro and iPhone 15 models, which have a lower prompt decoding rate, exhibit a higher standard deviation compared to others. However, except for the iPhone 15 Pro Max, the highest prompt decoding rates observed in the test set for all devices are very close to each other. Only the A17 Pro chip with a 6-core GPU shows a major performance boost.

\subsection{Inference Performance}
Inference (single-token decoding) refers to the LLM’s autoregressive prediction of the next token based on all the preceding tokens it has seen so far. The inference performance, depicted in Figure.\ref{fig:fig3}, refers to the same mechanism as prompt decoding. However, beyond the initial prompt decoding, inferring the next token requires decoding its logits one token at a time. This is because each new token depends on the previous one, which cannot be known in advance. In other words, evaluating the prompt is much faster in terms of tokens per second since the LLM can batch process the prompt tokens. Inference takes up the vast majority of total inference time and thus determines the fluency of the interaction between user and LLM and can be conceptualized as one of the essential metrics to evaluate on-device LLM performance.

Similar to prompt decoding, inference also runs on the GPU. However, contrary to the results of prompt decoding, our tests showed that the iPhone 15 Pro Max had the lowest inference rate. When the tests were repeated with a secondary iPhone 15 Pro Max, this result did not change. Unfortunately, we were unable to find a clear cause. Potential reasons for this mismatch between expected and actual performance could lie in changes to the GPU architecture or possibly a regression in the driver or even Metal. It could also be that the existing Metal shaders used are not optimized for underlying hardware changes. These tenuous hypotheses are supported by the observation that the non-Pro version of the iPhone 15 exhibits performance characteristics very similar to those of the iPhone 14 Pro models because it has the same chip; the slight difference can be explained by possible changes in the thermal dissipation properties of the new chassis of the phone, for example. Further investigation as well as rerunning the benchmark suite on future iOS versions are warranted. Besides that, results for all other iPhones we obtained were consistent with expectations. 

\section{Conclusion}
In this paper we discussed the need for on-device LLMs by addressing the requirements for safety, security, availability, and latency in AI applications. We then reviewed methods proposed in the literature for model compression, aiming to fit state-of-the-art LLMs on smartphones. Using existing models and frameworks from the literature, we developed a testbed to evaluate the on-device inference capabilities. Then, we investigated and evaluated the energy impacts of our test-suite. Finally, we evaluated the sampling, prompt decoding, and inference rates of different iPhones on a quantized 7B model. Our findings indicate that while recent iPhone generations possess the hardware capacity to run on-device LLMs, achieving sustained performance requires further advancements in power management and system integration. Future research should focus on optimizing these aspects to fully harness the potential of on-device LLMs. Additionally, we plan to explore ways to better compress and accelerate LLMs while minimizing output quality loss, as future work. 

\bibliographystyle{unsrt}  
\bibliography{references}

\begin{thebibliography}{10}

\bibitem{open-llm-leaderboard}
Edward Beeching, Clementine Fourrier, Nathan Habib, Sheon Han, Nathan Lambert, Nazneen Rajani, Omar Sanseviero, Lewis Tunstall, and Thomas Wolf.
\newblock Open llm leaderboard.
\newblock \url{https://huggingface.co/spaces/HuggingFaceH4/open_llm_leaderboard}, 2023.

\bibitem{marin_serverless_2022}
Eduard Marin, Diego Perino, and Roberto Di~Pietro.
\newblock Serverless computing: a security perspective.
\newblock {\em Journal of Cloud Computing}, 11(1):69, October 2022.

\bibitem{hinton_distilling_2015}
Geoffrey Hinton, Oriol Vinyals, and Jeff Dean.
\newblock Distilling the {Knowledge} in a {Neural} {Network}, March 2015.
\newblock arXiv:1503.02531 [cs, stat].

\bibitem{kim_sequence-level_2016}
Yoon Kim and Alexander~M. Rush.
\newblock Sequence-{Level} {Knowledge} {Distillation}, September 2016.
\newblock arXiv:1606.07947 [cs].

\bibitem{huang_-context_2022}
Yukun Huang, Yanda Chen, Zhou Yu, and Kathleen McKeown.
\newblock In-context {Learning} {Distillation}: {Transferring} {Few}-shot {Learning} {Ability} of {Pre}-trained {Language} {Models}, December 2022.
\newblock arXiv:2212.10670 [cs].

\bibitem{magister_teaching_2023}
Lucie~Charlotte Magister, Jonathan Mallinson, Jakub Adamek, Eric Malmi, and Aliaksei Severyn.
\newblock Teaching {Small} {Language} {Models} to {Reason}, June 2023.
\newblock arXiv:2212.08410 [cs].

\bibitem{gu_knowledge_2023}
Yuxian Gu, Li~Dong, Furu Wei, and Minlie Huang.
\newblock Knowledge {Distillation} of {Large} {Language} {Models}, June 2023.
\newblock arXiv:2306.08543 [cs].

\bibitem{zhu_survey_2023}
Xunyu Zhu, Jian Li, Yong Liu, Can Ma, and Weiping Wang.
\newblock A {Survey} on {Model} {Compression} for {Large} {Language} {Models}, September 2023.
\newblock arXiv:2308.07633 [cs].

\bibitem{frantar_sparsegpt_2023}
Elias Frantar and Dan Alistarh.
\newblock {SparseGPT}: {Massive} {Language} {Models} {Can} {Be} {Accurately} {Pruned} in {One}-{Shot}, March 2023.
\newblock arXiv:2301.00774 [cs].

\bibitem{sun_simple_2023}
Mingjie Sun, Zhuang Liu, Anna Bair, and J.~Zico Kolter.
\newblock A {Simple} and {Effective} {Pruning} {Approach} for {Large} {Language} {Models}, October 2023.
\newblock arXiv:2306.11695 [cs].

\bibitem{noune_8-bit_2022}
Badreddine Noune, Philip Jones, Daniel Justus, Dominic Masters, and Carlo Luschi.
\newblock 8-bit {Numerical} {Formats} for {Deep} {Neural} {Networks}, June 2022.
\newblock arXiv:2206.02915 [cs].

\bibitem{dettmers_llmint8_2022}
Tim Dettmers, Mike Lewis, Younes Belkada, and Luke Zettlemoyer.
\newblock {LLM}.int8(): 8-bit {Matrix} {Multiplication} for {Transformers} at {Scale}, November 2022.
\newblock arXiv:2208.07339 [cs].

\bibitem{shao_omniquant_2023}
Wenqi Shao, Mengzhao Chen, Zhaoyang Zhang, Peng Xu, Lirui Zhao, Zhiqian Li, Kaipeng Zhang, Peng Gao, Yu~Qiao, and Ping Luo.
\newblock {OmniQuant}: {Omnidirectionally} {Calibrated} {Quantization} for {Large} {Language} {Models}, August 2023.
\newblock arXiv:2308.13137 [cs].

\bibitem{mathur2023orcamini}
Pankaj Mathur.
\newblock orca mini v3 7b: An explain tuned llama2-7b model, 2023.

\bibitem{touvron2023llama}
Hugo Touvron, Louis Martin, Kevin Stone, Peter Albert, Amjad Almahairi, Yasmine Babaei, Nikolay Bashlykov, Soumya Batra, Prajjwal Bhargava, Shruti Bhosale, Dan Bikel, Lukas Blecher, Cristian~Canton Ferrer, Moya Chen, Guillem Cucurull, David Esiobu, Jude Fernandes, Jeremy Fu, Wenyin Fu, Brian Fuller, Cynthia Gao, Vedanuj Goswami, Naman Goyal, Anthony Hartshorn, Saghar Hosseini, Rui Hou, Hakan Inan, Marcin Kardas, Viktor Kerkez, Madian Khabsa, Isabel Kloumann, Artem Korenev, Punit~Singh Koura, Marie-Anne Lachaux, Thibaut Lavril, Jenya Lee, Diana Liskovich, Yinghai Lu, Yuning Mao, Xavier Martinet, Todor Mihaylov, Pushkar Mishra, Igor Molybog, Yixin Nie, Andrew Poulton, Jeremy Reizenstein, Rashi Rungta, Kalyan Saladi, Alan Schelten, Ruan Silva, Eric~Michael Smith, Ranjan Subramanian, Xiaoqing~Ellen Tan, Binh Tang, Ross Taylor, Adina Williams, Jian~Xiang Kuan, Puxin Xu, Zheng Yan, Iliyan Zarov, Yuchen Zhang, Angela Fan, Melanie Kambadur, Sharan Narang, Aurelien Rodriguez, Robert Stojnic, Sergey Edunov, and Thomas
  Scialom.
\newblock Llama 2: Open foundation and fine-tuned chat models, 2023.

\bibitem{mukherjee2023orca}
Subhabrata Mukherjee, Arindam Mitra, Ganesh Jawahar, Sahaj Agarwal, Hamid Palangi, and Ahmed Awadallah.
\newblock Orca: Progressive learning from complex explanation traces of gpt-4, 2023.

\bibitem{llama.cpp}
Georgi Gergenov.
\newblock llama.cpp b1407.
\newblock \url{https://github.com/ggergenov/llama.cpp}, 2023.

\end{thebibliography}

\end{document}